\documentclass[conference]{IEEEtran}
\IEEEoverridecommandlockouts

\usepackage{cite}
\usepackage{amsmath,amssymb,amsfonts}
\usepackage{algorithm}
\usepackage{algorithmic}
\usepackage[dvipdfmx]{graphicx}
\usepackage[dvipdfmx]{color}
\usepackage{textcomp}
\usepackage{xcolor}
\usepackage{subcaption}
\usepackage{url}
\usepackage{siunitx}

\def\BibTeX{{\rm B\kern-.05em{\sc i\kern-.025em b}\kern-.08em
    T\kern-.1667em\lower.7ex\hbox{E}\kern-.125emX}}

\newcommand{\sig}[1]{\textsuperscript{%
  \ifdim #1 pt < 0.001pt ***%
  \else\ifdim #1 pt < 0.01pt **%
  \else\ifdim #1 pt < 0.05pt *%
  \else\fi\fi\fi}}

\begin{document}

\title{A Memetic Algorithm based on Variational Autoencoder for Black-Box Discrete Optimization with Epistasis among Parameters\\
}

\author{\IEEEauthorblockN{1\textsuperscript{st} Aoi Kato}
\IEEEauthorblockA{\textit{School of Engineering} \\
\textit{Tokyo Institute of Technology}\\
Yokohama, Japan \\
kato.a@ic.comp.isct.ac.jp
}
\and
\IEEEauthorblockN{2\textsuperscript{nd} Kenta Kojima}
\IEEEauthorblockA{\textit{School of Computing} \\
\textit{Tokyo Institute of Technology}\\
Yokohama, Japan \\
kenta@ic.c.titech.ac.jp}
\and
\IEEEauthorblockN{3\textsuperscript{rd} Masahiro Nomura}
\IEEEauthorblockA{\textit{School of Computing} \\
\textit{Tokyo Institute of Technology}\\
Yokohama, Japan \\
nomura@comp.isct.ac.jp}
\and
\IEEEauthorblockN{4\textsuperscript{rd} Isao Ono}
\IEEEauthorblockA{\textit{School of Computing} \\
\textit{Tokyo Institute of Technology}\\
Yokohama, Japan \\
isao@comp.isct.ac.jp}
\and
}

\IEEEoverridecommandlockouts
\IEEEpubid{\makebox[\columnwidth]{979-8-3315-3431-8/25/\$31.00~\copyright2025 IEEE \hfill}
\hspace{\columnsep}\makebox[\columnwidth]{}}

\maketitle

\IEEEpubidadjcol

\begin{abstract}
Black-box discrete optimization (BB-DO) problems arise in many real-world applications, such as neural architecture search and mathematical model estimation. A key challenge in BB-DO is epistasis among parameters where multiple variables must be modified simultaneously to effectively improve the objective function.
Estimation of Distribution Algorithms (EDAs) provide a powerful framework for tackling BB-DO problems.
In particular, an EDA leveraging a Variational Autoencoder (VAE) has demonstrated strong performance on relatively low-dimensional problems with epistasis while reducing computational cost.
Meanwhile, evolutionary algorithms such as DSMGA-II and P3, which integrate bit-flip-based local search with linkage learning, have shown excellent performance on high-dimensional problems.
In this study, we propose a new memetic algorithm that combines VAE-based sampling with local search.
The proposed method inherits the strengths of both VAE-based EDAs and local search-based approaches: it effectively handles high-dimensional problems with epistasis among parameters \emph{without} incurring excessive computational overhead.
Experiments on NK landscapes—a challenging benchmark for BB-DO involving epistasis among parameters—demonstrate that our method outperforms state-of-the-art VAE-based EDA methods, as well as leading approaches such as P3 and DSMGA-II.

\end{abstract}


\section{Introduction}
\label{sec:intro}

Black-box discrete optimization (BB-DO) plays a critical role in a variety of real-world applications, including neural architecture search~\cite{NAS,elsken2019neural} and mathematical model estimation~\cite{SR}.
Unlike continuous optimization, BB-DO deals with discrete search spaces where gradient-based methods are often inapplicable, making heuristic and population-based algorithms a popular choice.
A key challenge in BB-DO is epistasis among parameters which means that modifying a single variable independently may not lead to meaningful improvements in the objective function.

VAE-EDA-Q~\cite{VAE-EDA-Q} is a state-of-the-art Estimation of Distribution Algorithms (EDAs) \cite{EDAs2002,EDAs2011} that employs a Variational Autoencoder (VAE) as its probabilistic model to curb premature convergence and capture complex epistasis among parameters, thereby reducing computational cost.
It has attracted attention as a promising remedy for the scalability issue—namely, the exponential increase in computational effort with growing problem size and dependency complexity—that plagues the Bayesian Optimization Algorithm (BOA)~\cite{BOA}, which represents problem structure by learning a Bayesian network. EDAs form a broad framework that builds probabilistic models of promising solutions to generate new candidates, spanning methods that use univariate models such as PBIL \cite{PBIL} and UMDA \cite{UMDA} to multivariate models capable of representing epistasis among parameters, including MIMIC~\cite{MIMIC} and BOA~\cite{BOA}.
In particular, BOA-based methods face prohibitive model-construction cost as dimensionality increases \cite{EDAs2011}.
Although VAE-EDA-Q alleviates this limitation, it has so far been evaluated mainly on relatively low-dimensional problems and may still face challenges in scaling efficiently to higher dimensions.

On the other hand, evolutionary algorithms (EAs) that integrate bit-flip-based local search with linkage learning, such as DSMGA-II~\cite{DSMGA-II}, Parameter-less Population Pyramid (P3)~\cite{P3} and P3-FIHCwLL~\cite{FIHCwLL}, have shown impressive performance compared to existing linkage learning methods such as LTGA~\cite{LTGA} and multi-start local search~\cite{P3} on high-dimensional BB-DO problems.
These methods leverage bit-flip-based local search strategies to efficiently explore the search space while identifying and exploiting problem structure.
Despite their effectiveness, they can be computationally expensive due to the iterative nature of local search.

In this study, we propose a novel memetic algorithm that synergistically combines VAE-based sampling with local search to address the challenges of high-dimensional BB-DO problems with epistasis among parameters.
Our approach leverages the expressive power of VAEs to model dependencies among variables while integrating local search to refine solutions efficiently.
By combining these strengths, our method achieves high optimization performance without incurring excessive computational overhead.
Specifically, the proposed method has the following ideas:
\begin{itemize}
    \item As in VAE-EDA-Q, employing a VAE for offspring generation can leverage the epistasis among parameters inherent in the population, while also helping suppress the increase in computation time for higher-dimensional problems.
    \item By applying a bit-flip-based local search to the EDA-generated individuals, we can discover higher-fitness individuals that may not be found solely via sampling from the VAE.
\end{itemize}

To evaluate the effectiveness of our approach, we conduct experiments on NK landscapes~\cite{kauffman1989nk}, a well-established benchmark for BB-DO that explicitly incorporates epistasis among parameters, with various epistasis strengths.
Specifically, we set the problem size to $N=300, 1000$ to assess the scalability of our method in high-dimensional settings.
Our results show that the proposed method outperforms state-of-the-art BB-DO methods such as VAE-EDA-Q, P3, and DSMGA-II.
These findings highlight the potential of hybrid approaches that blend probabilistic modeling with heuristic search, offering a promising direction for tackling challenging BB-DO problems in practical settings.

\section{Proposed Method}\label{sec:proposed}
In this section, we propose a memetic algorithm that generates offspring via sampling from a VAE and applies a bit-flip-based local search to improve these offspring.
Whereas conventional VAE-EDA-based approaches generate new individuals solely by sampling from the VAE, the proposed method utilizes those sampled individuals as initial solutions for local search.
This process yields individuals with better fitness values and greater diversity than those in the current population.
Additionally, to mitigate premature convergence in the population, the method generates multiple offspring from a single parent and excludes previously discovered individuals, thereby maintaining population diversity.
\par
\begin{figure}[tb]
    \centering
    \includegraphics[width=\hsize]{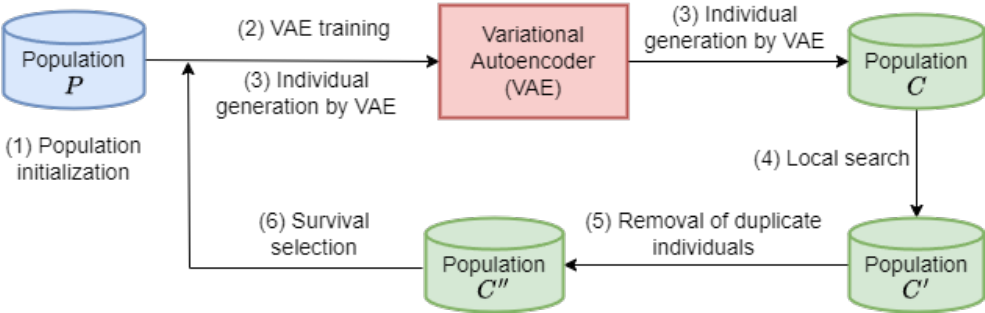}
    \caption{Flow of the proposed method: (1) Population initialization, (2) VAE training, (3) Individual generation by VAE, (4) Local search, (5) Removal of duplicate individuals, (6) Survival selection.}
    \label{fig:proposed-flow}
\end{figure}

The proposed method consists of the following six steps, as shown in Fig.~\ref{fig:proposed-flow}:
\begin{enumerate}
    \item Uniformly sample $\lambda$ individuals from the search space and apply a bit-flip-based local search to them, creating the initial population $P$.
    \item Train the VAE using the population $P$.
    \item Generate an offspring population $C$ by sampling from the trained VAE.
    \item Apply a bit-flip-based local search to the offspring population $C$, producing $C'$.
    \item Remove duplicate individuals in $C'$ that have already been generated, yielding $C''$.
    \item Select the next generation of \(\lambda\) individuals based on fitness from \( P \cup C'' \). If the termination condition is not met, return to Step~2.
\end{enumerate}
\par

\subsection{Training VAE using the population}\label{sec:learning}
In the proposed method, the entire population is used as training data in each generation to construct the VAE, thereby representing the search space in a latent space.
Hyperparameters for VAE construction are chosen to ensure that the model accurately reconstructs the population's individuals in the latent space.
Furthermore, to stabilize training, batch normalization \cite{batchnorm} is applied after each fully connected layer.
The detailed settings are provided in Section~\ref{sec:exp_setting}.

\subsection{Generating Individuals via VAE}\label{sec:generation}
To prevent the population from converging prematurely, the process of generating offspring from a single parent using the VAE is repeated \(n_\text{vae}\) times, where $n_\text{vae}$ is a user parameter.
Duplicate offspring are detected using a hash list that keeps track of all individuals previously generated by the VAE and local search.
Since the decoder outputs individuals as probabilities in the range \([0,1]\), each bit is discretized to one if its probability is \(0.5\) or higher, and zero otherwise.

\subsection{Local Search}\label{sec:local_search}
The newly generated individuals from the VAE serve as initial solutions for local search.
We employ the First Improvement Hill Climber (FIHC) used in P3 \cite{P3} as the local search method.
The procedure for FIHC is as follows.
First, the bit positions in the individual are randomized.
Then, in that randomized order, each bit is flipped in turn.
If flipping a bit improves the objective function value, the change is retained; otherwise, the bit is reverted to its original state.
If no improvement is found after trying all bits, the local search is terminated.
To avoid redundant evaluations, the bit positions flipped since the last change are recorded.

\subsection{Survival Selection}\label{sec:survivor_selection}
After local search, duplicate offspring are removed, so that offspring population contains only one copy of each unique individual.
Then, \(\lambda\) individuals are selected from the combined set of the current population and the offspring population based on their fitness values, where \(\lambda\) is the population size.

\section{Experiments}\label{sec:experiments}
In this section, we compare the performance of the proposed method with that of DSMGA-II, P3, VAE-EDA-Q, and a multi-start local search method on benchmark problems that involve epistasis among parameters.

\subsection{Comparison Methods}
DSMGA-II \cite{DSMGA-II} is an evolutionary algorithm that constructs a linkage model based on the dependency structure of variables represented by a Dependency Structure Matrix (DSM), and then performs a search using two types of crossover.
P3 \cite{P3} is a method that employs hierarchical population management and linkage-aware crossover. Both methods have demonstrated strong performance on high-dimensional NK Landscapes.
VAE-EDA-Q \cite{VAE-EDA-Q} is a VAE-based EDA designed to mitigate premature convergence.
The multi-start local search (MSLS) method repeatedly applies the First Improvement Hill Climber, used in the proposed method, to randomly generating individuals until the termination condition is met.

\subsection{Benchmark problems}
In this study, we use the NK Landscapes~\cite{kauffman1989nk}, a representative benchmark for black-box discrete optimization. In NK Landscapes, the fitness contribution of each of the \( N \) variables depends on \( k \) other variables, where the parameter \( k \) governs the complexity of the problem.
Individuals are represented by an \(N\)-dimensional binary vector \(\mathbf{x} = (x_1, \ldots, x_N)\), and the objective function is defined as:
\begin{align}
    f(\mathbf{x}) = \frac{1}{N} \sum_{i=1}^{N} f_i(x_i, x_{i_1}, \ldots, x_{i_k}),
\end{align}
where \( f_i \) is a subfunction based on variable \( x_i \) and the \( k \) variables \(\{x_{i_1}, \ldots, x_{i_k}\}\) on which it depends. Each value of \( f_i \) for a combination \((x_i, x_{i_1}, \ldots, x_{i_k})\) is generated by assigning a uniform random number \( U(0,1) \). The specific sets of variables upon which each variable depends are generated randomly.
\par
In this section, we consider \( N=\{300, 1000\} \) and \( k=\{4, 5, 6\} \). For each combination \((N,k)\), a single problem instance is generated using a fixed random seed.

\subsection{Evaluation Criteria}
We employ the average best fitness obtained at the end of optimization across 50 trials as the evaluation criteria to compare the performance of each method.
For DSMGA-II, the population size that achieves the highest average best fitness in preliminary experiments is used for the 50-trial runs reported here. The population sizes of the proposed method and VAE-EDA-Q are left untuned, and the remaining algorithms require no tuning at all.

\subsection{Experimental Settings}\label{sec:exp_setting}
The hyperparameters of the proposed method are set as follows. The population size is \(\lambda = N\), and the number of offspring generated per parent \( n_\text{vae} \) is set to 10.
For the VAE,
we use one fully connected layer each in the encoder and decoder,
4,096 nodes in the hidden layer,
and a latent space dimension \( d_\text{latent} = 32 \).
The encoder and decoder each consist of an input layer, a fully connected layer, a batch normalization layer, and an activation function in that order. We use ReLU as the activation function after each batch normalization layer, and a sigmoid function in the output layer of the decoder.
For VAE training, the batch size \( n_\text{batch} \) is 64, the number of epochs \( n_\text{epochs} \) is 500, the learning rate is \(\alpha = 0.001\), and we use the Nadam \cite{nadam} optimizer.
VAE-EDA-Q is set up with the same parameters as the proposed method.
For DSMGA-II, the population sizes we consider are \(\lambda = \{100, 200, 300, 400, 500\}\) for \(N=300\) and \(\lambda = \{750, 1000, 1250, 1500\}\) for \(N=1000\). We choose the population size that yields the highest average best fitness in the final generation.

The optimization terminates when the number of function evaluations reaches a given maximum number of function evaluations given by a user. The maximum number of function evaluations is \(3 \times 10^7\) for \(N=300\) and \(3 \times 10^8\) for \(N=1000\).

\subsection{Implementation}
This section describes the implementation details of each method.
For the proposed method, the VAE is implemented in Python using PyTorch \cite{pytorchLibrary}.
For DSMGA-II, we use the C++ implementation publicly available at \url{https://teilab.ee.ntu.edu.tw/resources/}.
For P3, we use the implementation provided at \url{https://github.com/brianwgoldman/Parameter-less_Population_Pyramid}.
For VAE-EDA-Q, we use the implementation provided at \url{https://github.com/sourodeep/vaeedaq}. However, since the provided code does not work because of missing methods, we corrected them based on the original VAE-EDA-Q paper.

\subsection{Results}
Table~\ref{tab:results1} shows the mean and standard deviation of the best fitness obtained after 50 independent trials by each method. \textit{Diff} means the difference between the mean obtained by the proposed method and that of each conventional method with significance markers obtained by Dunn's post-hoc test with Holm correction (\textsuperscript{*}\,$p<0.05$, \textsuperscript{**}\,$p<0.01$, \textsuperscript{***}\,$p<0.001$). Note that, if \textit{Diff} is positive, the proposed method outperforms the conventional method.
As shown in Table~\ref{tab:results1}, the proposed method achieves the highest average fitness in every \((N,k)\) setting.
For DSMGA-II, VAE-EDA-Q, and MSLS, every difference is highly significant (\textsuperscript{***}).
Against P3 the advantage is also significant in five of the six cases (three \textsuperscript{***} and two \textsuperscript{**}); only for the configuration \((N{=}1000,k{=}4)\),  the difference is not significant (no star).
The results suggest that the proposed method is effective.

\begin{table*}[tb]
	\centering
    \caption{Mean and standard deviation of the best fitness values at the end of optimization across 50 trials on the NK Landscapes. Higher values indicate better results. Bold font denotes the highest mean fitness in each row. "Diff" represents the difference from the proposed method's mean fitness, where a positive value implies that the proposed method found better solutions. Asterisks ($^*$/$^{**}$/$^{***}$) denote significance at the 0.05/0.01/0.001 levels after Dunn tests with Holm correction.}
	\label{tab:results1}
	\begin{tabular}{rrrrrrr}
		\hline
		$N$ & $k$ & Proposed & DSMGA-II & Diff (vs. DSMGA-II) & P3 & Diff (vs. P3)\\ \hline
		 $300$ & $4$ & $\mathbf{0.7817 \pm 0.0025}$ & $0.7652 \pm 0.0026$ & $+0.0162\sig{0.0000}$ & $0.7711 \pm 0.0035$ & $+0.0103\sig{0.0030}$\\
		 $300$ & $5$ & $\mathbf{0.7882 \pm 0.0036}$ & $0.7641 \pm 0.0051$ & $+0.0229\sig{0.0000}$ & $0.7740 \pm 0.0040$ & $+0.0130\sig{0.0005}$\\
		 $300$ & $6$ & $\mathbf{0.7832 \pm 0.0035}$ & $0.7568 \pm 0.0027$ & $+0.0257\sig{0.0000}$ & $0.7638 \pm 0.0036$ & $+0.0187\sig{0.0001}$\\
        $1000$ & $4$ & $\mathbf{0.7684 \pm 0.0012}$ & $0.7540 \pm 0.0023$ & $+0.0112\sig{0.0000}$ & $0.7610 \pm 0.0025$ & $+0.0042\sig{0.1231}$\\
		$1000$ & $5$ & $\mathbf{0.7694 \pm 0.0012}$ & $0.7513 \pm 0.0022$ & $+0.0183\sig{0.0000}$ & $0.7598 \pm 0.0025$ & $+0.0098\sig{0.0010}$\\
		$1000$ & $6$ & $\mathbf{0.7706 \pm 0.0033}$ & $0.7506 \pm 0.0022$ & $+0.0180\sig{0.0000}$ & $0.7552 \pm 0.0020$ & $+0.0134\sig{0.0008}$\\\hline
        $N$ & $k$ & Proposed & VAE-EDA-Q & Diff (vs. VAE-EDA-Q) & Multi-Start Local Search (MSLS) & Diff (vs. MSLS) \\ \hline
		 $300$ & $4$ & $\mathbf{0.7817 \pm 0.0025}$ & $0.5359 \pm 0.0042$ & $+0.2458\sig{0.0000}$ & $0.7611 \pm 0.0030$ & $+0.0203\sig{0.0000}$ \\
		 $300$ & $5$ & $\mathbf{0.7882 \pm 0.0036}$ & $0.5411 \pm 0.0035$ & $+0.2471\sig{0.0000}$ & $0.7633 \pm 0.0028$ & $+0.0237\sig{0.0000}$ \\
		 $300$ & $6$ & $\mathbf{0.7832 \pm 0.0035}$ & $0.5391 \pm 0.0051$ & $+0.2441\sig{0.0000}$ & $0.7609 \pm 0.0026$ & $+0.0314\sig{0.0000}$ \\
        $1000$ & $4$ & $\mathbf{0.7684 \pm 0.0012}$ & $0.5618 \pm 0.0088$ & $+0.2066\sig{0.0000}$ & $0.7460 \pm 0.0014$ & $+0.0192\sig{0.0000}$ \\
		$1000$ & $5$ & $\mathbf{0.7694 \pm 0.0012}$ & $0.5658 \pm 0.0084$ & $+0.2036\sig{0.0000}$ & $0.7486 \pm 0.0016$ & $+0.0210\sig{0.0000}$ \\
		$1000$ & $6$ & $\mathbf{0.7706 \pm 0.0033}$ & $0.5646 \pm 0.0104$ & $+0.2060\sig{0.0000}$ & $0.7493 \pm 0.0008$ & $+0.0193\sig{0.0000}$ \\\hline
	\end{tabular}
\end{table*}

\section{Conclusion}\label{sec:conclusion}
In this study, we proposed a memetic algorithm for discrete black-box function optimization problems with epistasis among parameters.
The proposed method employs VAE to generate offspring, and a bit-flip-based local search to improve the offspring.
In experiments using NK Landscapes with \(N=300\) and \(1000\), and \(k=4, 5, 6\), the proposed method outperformed DSMGA-II, VAE-EDA-Q, and a multi-start local search method based on bit-flipping. The proposed method showed better performance than P3 on almost all the settings except when $N=1000$ and $k=4$. 
\par
In future work, we have to investigate the reason the proposed method did not outperform P3 when $N=1000$ and $k=4$. Also, we would like to propose a way of dynamically adjusting the parameter \(n_\text{vae}\), which is the number of offspring generated per parent, in order to enable to discover better solutions with fewer evaluations. Furthermore, we have a plan to extend the proposed method to continuous and mixed-integer optimization problems.


\bibliography{biblio}
\bibliographystyle{IEEEtran}

\end{document}